\documentclass{article}

\usepackage{microtype}
\usepackage{graphicx}
\usepackage{subfigure}
\usepackage{amsmath,amssymb,amsfonts}
\usepackage{xurl}
\usepackage{algorithmic}
\usepackage{enumerate}
\usepackage{textcomp}
\usepackage{xcolor}
\usepackage{url}
\usepackage{units}
\usepackage{csquotes}
\usepackage{booktabs} 

\usepackage{hyperref}

\newcommand{\set}[1]{\ensuremath\mathcal{#1}}
\newcommand{\vect}[1]{\ensuremath\mathbf{#1}}
\newcommand{\card}{\ensuremath\vect{c}}
\newcommand{\mycards}{\ensuremath\set{C}}
\newcommand{\pack}{\ensuremath\set{P}}
\newcommand{\pref}{\succ}

\usepackage[accepted]{icml2021}

\usepackage{comment}
\usepackage{todonotes}

\icmltitlerunning{Preference Ranking for Set Addition Problems}

\begin{document}

\twocolumn[
\icmltitle{A Comparison of Contextual and Non-Contextual \\Preference Ranking for Set Addition Problems}



\icmlsetsymbol{equal}{*}

\begin{icmlauthorlist}
\icmlauthor{Timo Bertram}{JKU}
\icmlauthor{Johannes F\"urnkranz}{JKU}
\icmlauthor{Martin M\"uller}{CA}
\end{icmlauthorlist}

\icmlaffiliation{JKU}{Dept.\ of Computer Science,
Johannes-Kepler Universit\"at,
Linz, Austria}
\icmlaffiliation{CA}{Dept.\ of Computing Science,
University of Alberta,
Edmonton, Canada}

\icmlkeywords{Machine Learning, ICML}

\icmlcorrespondingauthor{Timo Bertram}{tbertram@faw.jku.at}

\vskip 0.3in
]



\printAffiliationsAndNotice{}  

\begin{abstract}
In this paper, we study the problem of evaluating the addition of elements to a set. This problem is difficult, because it can, in the general case, not be reduced to unconditional preferences between the choices. Therefore, we model preferences based on the context of the decision. We discuss and compare two different Siamese network architectures for this task: a twin network that compares the two sets resulting after the addition, and a triplet network that models the contribution of each candidate to the existing set.
We evaluate the two settings on a real-world task; learning human card preferences for deck building in the collectible card game \enquote{Magic: The Gathering}. We show that the triplet approach achieves a better result than the twin network and that both outperform previous results on this task.
\end{abstract}

\section{Introduction}
Set addition problems can be commonly found in many applications. The problem is to evaluate which of several possible candidates is the best addition to an existing set, such that the resulting set achieves a high evaluation according to a latent set evaluation function.Examples include adding cards to a player's deck, adding players to a football team, or buying stocks to complement an existing portfolio. Naturally, the evaluation of such items depends on the elements that are already in the set. For example, a mediocre goalkeeper may be the better addition to a team of excellent field players than the best striker, and the evaluation of a promising high-risk share may depend on the risk profile of the stocks you are already holding.

In this paper, we look at this problem in a preference learning context \citep{plbook}: we assume that we are given training information that specifies which of two possible choices is preferential over the other. Based on this problem formulation, we study and compare two different learning schemes based on Siamese neural networks. The first is a classical preference learning setting, where the learner is trained to predict which of the two sets resulting from the addition is preferable. As a second variant, we consider a setting that can directly model which of the two additions fits the context better. As such, the former implicitly models the context of the decision by comparing the two resulting sets, while the latter models the context explicitly as a separate input to the learner. 

We formally define contextual preferences and the set addition problem in Section~\ref{sec:CPR}, show two Siamese neural network approaches to this problem in Section~\ref{sec:networks}, and evaluate and compare them for the task of deck-building in the collectible card game \emph{Magic: The Gathering (MTG)}.
To train and evaluate our method, we use a dataset of sequential expert deck-building decisions, which provides information about which selections the human experts preferred over others and allows us to compare the two preference-based methods on this data. These experiments and their results are presented in Sections~\ref{sec:experiments} and~\ref{sec:results}, followed by some conclusions. This work builds open our previous research for the Contextual Preference Ranking framework \cite{CPR}.

\section{Contextual Preferences and the Set Addition Problem}
\label{sec:CPR}

As described above, we are concerned with problems where we have to evaluate the addition of an item to an existing set of items, such as adding players to a sports team, buying stocks for a portfolio, putting products in a shopping basket, or adding cards to a player's deck. In the general case, this problem is hard, because the value of the items that are added changes depending on the set it is part of. In many cases, items may have different, hidden properties, and the evaluation of the set depends on properties that are covered by the individual items. While each item a value of its own, and some are better than others, the overall value of a set does not equal the sum of values of items. It crucially depends on the overall composition.

Formally, this \emph{set addition problem} can be represented as follows:
Given a set of items $\mycards$ as the \emph{context}, and a set of items $\pack$ that represent the current possible choices, select the item $\card^*$ in  $\pack$ which is the best addition to  $\mycards$. Let us assume an (unknown) utility function $u(.)$ which returns an evaluation for a given set of items, then
\begin{equation}
    \card^{*} = \arg \max_{\card \in \pack} u(\mycards \cup \{\card\})
\end{equation}
The learning problem is to learn the function $u(.)$ from a set of example decisions. The training information can be given in various ways. In this paper, we assume a preference-based formulation: we do not have access to a direct evaluation of a set, but are given pairwise comparisons between them. In particular, we assume that we have access to a set of \emph{contextual preferences} of the form 
\begin{equation}
\label{eq:CjoverCk}
        \left(\card_j \pref \card_k \mid \mycards\right)
\end{equation}
which means that item $\card_j$ is a better addition to the context $\mycards$ than item $\card_k$. 
In our set-based setting, this is equivalent to  unconditional preferences of the form
\begin{equation}
\label{eq:Unconditional Preferences}
        \left(\mycards \cup \{\card_j\}\right) \pref \left(\mycards \cup \{\card_k\} \right).
\end{equation}
The main contribution of this paper is a case study that translates these two formulations into corresponding neural network architectures and compares them on learning human preferences in a real-world card game. For decisions without a context, when the first item is added to the set, $\mycards = \emptyset$.  This special case of an empty context may be viewed as a comparison of the general, unconditional utility of two items $\card_j$ and $\card_k$.

While we will not tackle this in the current paper, this framework can in principle easily be generalized to using arbitrarily large sets  of items $\mycards_j$ and $\mycards_k$ for the context, i.e., for dealing with contextual preferences of the form
\begin{equation}
        \left(\mycards_j \pref \mycards_k \mid \mycards_i\right).
\end{equation}

\section{Learning Contextual Preferences with Siamese Networks}
\label{sec:networks}

In the following, we briefly describe how to use Siamese networks for preference learning from sets. While they are typically used on multiple examples of the same type, e.g. images, we employ them to allow comparisons of two items with a context by embedding both inputs as well as the context in a uniform representation space. 
To the best of our knowledge, this is a novel approach.

\subsection{Siamese networks for preference learning}
\label{sec:Siamese}

Siamese networks implement the idea that the same network is used multiple times in order to encode multiple items. The encodings are then compared and trained by a supervision signal \cite{b8}. A prototypical application of such networks is one-shot learning for image recognition \cite{b14,b9}.
Going back to \emph{comparison training} \citep{lig*Tesauro89b}, similar symmetrical architectures have also been used in preference learning, where the task is not to encode the similarity of objects but the preference between them.

A more traditional neural network approach to preference learning would compare two items by having both as a concatenated input to one network, which then outputs a single signal to model a preference. One problem with this is that reordering the inputs can lead to different results. While this can in practice be combated by training the network with random orderings, there is no guarantee that this fully eliminates the error. An order-dependent output is problematic and should not occur in practice. Siamese networks circumvent this problem by processing multiple inputs sequentially by the same network. This leads to a separate output for each input, called the \textbf{embedding} of the input.

\subsection{Unconditional Preferences}
\label{sec:RankNet}

\begin{figure}[t]
    \centering
    \includegraphics[width = 0.5\linewidth]{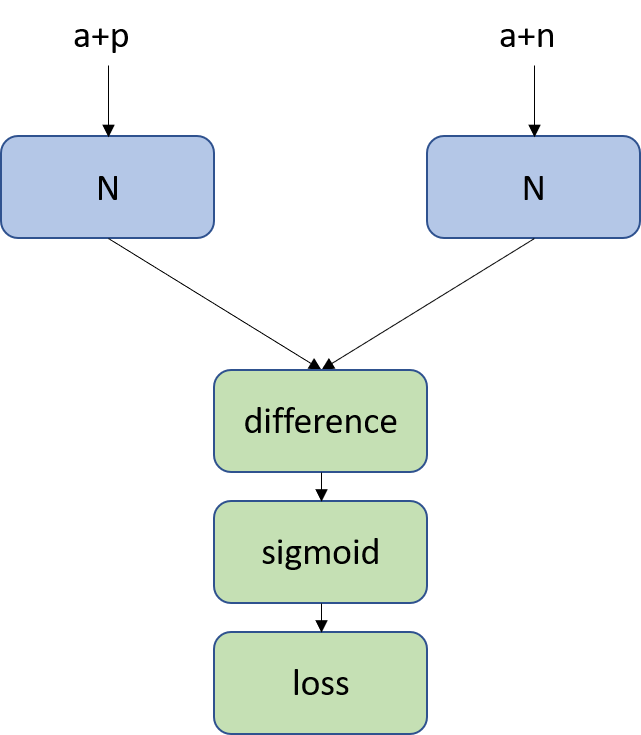}
    \caption{Training scheme for cross-entropy loss using a positive $\vect{p}$ and a negative example $\vect{n}$. The loss function aims to increase the evaluation of $\vect{p}$ over the one of $\vect{n}$. $N$ is the network that maps the two objects to an evaluation.}
    \label{fig:cross_entropy}
\end{figure}

One way to model the preference of which item to add to a set is to model the preferences over the resulting unions.
To use preferences of this type, one branch of the Siamese network encodes the preferred object $\vect{p}$, and the other branch the losing object $\vect{n}$. The two encodings are then compared by using their difference, as in Figure~\ref{fig:cross_entropy}.

For sets, we define $\vect{p} = \mycards \cup \{\card_j\}$ and $\mathbf{n} = \mycards \cup \{\card_k\}$, as shown in~\eqref{eq:Unconditional Preferences}. The output of the network is a single real value, which can be regarded as an evaluation $u(\mycards \cup \{\card_j\})$ of the set $\vect{p}$. The preference of one set over another is modeled by a higher evaluation $u(.)$. 

This setting corresponds to comparison training, which has been proposed by \citet{lig*Tesauro89b} in a game-playing context. For comparisons between two arbitrary items, the RankNet approach \cite{RankNet} uses a cross-entropy loss function and the sigmoid of the difference between the two evaluations.
We directly follow this method in our first set-based approach and will refer to it as \emph{RankNet}.

\subsection{Triplet Siamese Networks for Contextual Preference Ranking}
\label{sec:triple}

\begin{figure}[t]
    \centering
    \includegraphics[width = 0.72\linewidth]{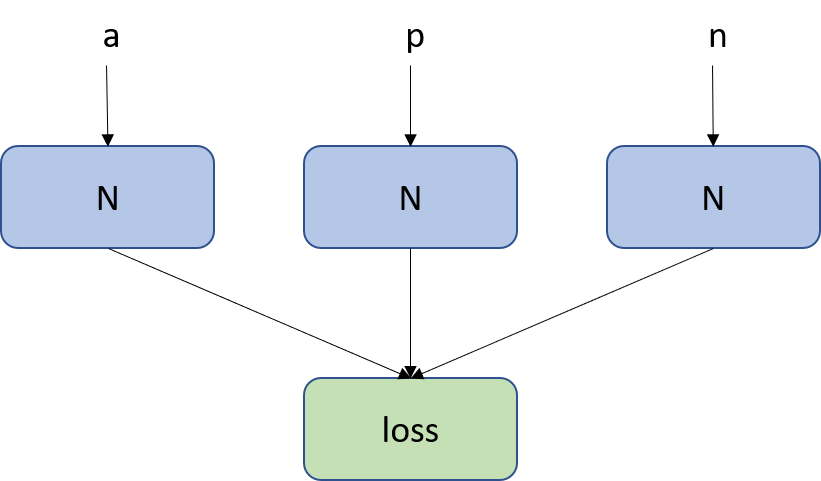}
    \caption{Training scheme for triplet loss using an anchor $\vect{a}$, a positive $\vect{p}$ and a negative example $\vect{n}$. $N$ is the network that maps each of the three objects into an embedding space. The loss function indicates whether $\vect{a}$ is closer to $\vect{p}$ or to $\vect{n}$. 
    }
    \label{fig:scheme}
\end{figure}

For directly using the contextual preferences of the form~\eqref{eq:CjoverCk}, we employ triplet Siamese networks, as shown in Figure~\ref{fig:siamese}. The key idea of this approach is to use an anchor ($\vect{a}$), a positive ($\vect{p}$) and a negative ($\vect{n}$) example. The anchor models the context, and the positive example $\vect{p}$ is preferred to the negative example $\vect{n}$ in this context, $\left(\vect{p}\pref \vect{n} \mid \vect{a}\right)$.

Such networks are trained with a triplet loss
\begin{equation}
\label{eq:triplet}
    L_{\textrm{triplet}}(\vect{a},\vect{p},\vect{n}) = \max\left(d(\vect{a},\vect{p})-d(\vect{a},\vect{n}) + m,0\right).
\end{equation}
The loss decreases with decreasing distance between $\vect{a}$ and $\vect{p}$, and with increasing distance between $\vect{a}$ and $\vect{n}$. This moves the embeddings of the anchor and the positive example closer together, and pushes the embedding between the anchor and the negative example apart. For this work, the Euclidian distance $d(\vect{x},\vect{y}) = ||\vect{x} - \vect{y}||_2$. is chosen as the distance metric between embeddings.
The \emph{margin} $m$ is a parameter of the loss function and controls how far embeddings are pushed away from each other. We used a margin of $m=1$. In preliminary experiments, the exact value of this parameter was not critical for the performance of the method. 
As an example, Siamese architectures can compare pictures of individuals and be trained to recognize whether two different images show the same person. In that case, the preference indicates which picture is more likely to show the same individual as the anchor.

We use them here in a slightly different, set-based setting. In our case, the anchor object $\vect{a}$ is the context set $\mycards$, which needs to be extended with one of two candidate extensions $\vect{p} = \card_j$ or $\vect{n} = \card_k$. The training information indicates that $\vect{p}$ is a better extension than $\vect{n}$. 
This is very different to asking whether $\vect{a}$ is more similar to $\vect{p}$ or $\vect{n}$. For example, card selection tasks seek cards that complement the deck, rather than duplicating the effect of similar cards picked earlier.

\subsection{Discussion}
At testing time, we do not need to query all possible pairwise comparisons of options, but can directly evaluate each option to formulate an overall ranking. In the case of Contextual Preference Ranking, this becomes possible because the resulting preferences are transitive w.r.t.\ to the given anchor set, i.e., 
\begin{equation*}
(\card_1 \pref \card_2 \mid \mycards)
\land
(\card_2 \pref \card_3 \mid \mycards)
\Rightarrow
(\card_1 \pref \card_3 \mid \mycards)
\end{equation*}
The reason for this is that all objects are embedded with the same embedding network $N$, which always outputs the same signal for the same input, regardless of the position of the item in the comparison. The same principle applies to unconditional preferences.



We view the adaptation of triplet Siamese networks to a set-based Contextual Preference Ranking framework as the main contribution of this work, as it introduces a new way of thinking about the Siamese triplet structure. Instead of comparing similar items, we train a preference of items based on a context. To our knowledge, we are the first to use triplet Siamese networks in such a way \cite{CPR}. This contextual preference of comparing $\vect{p}$ and $\vect{n}$ with context $\vect{a}$ also differs from trying to model the unconditional preferences $\vect{a}+\vect{p}$ and $\vect{a}+\vect{n}$. We want to emphasize the generality of this framework; it is applicable to model any kind of preference learning problem with a context.

\section{Experimental setup}
\label{sec:experiments}

The goal of our experimental evaluation is to compare the two different solutions for contextual preference problems described in Sections~\ref{sec:RankNet} and~\ref{sec:triple}. As a domain, we choose the problem of \textit{drafting}, or selecting cards, in the collectible card game \emph{Magic: The Gathering (MTG)}. 
We define the context $\mycards$ as the set of previously chosen cards of a player and train the networks with pairs of cards $\vect{p}$ and $\vect{n}$, where $\vect{p}$ was chosen by the player and $\vect{n}$ is another card that was available but not chosen. For Contextual Preference Ranking, we model that in the human expert's opinion, $\vect{p}$ fits better into the current set $\mycards$ than $\vect{n}$. For RankNet, we model that the set $\mycards 
\cup \{\vect{p}\}$ should receive a higher evaluation than $\mycards \cup \{\vect{n}\}$. Both approaches rank choices; CPR by distance to the anchor and RankNet by the evaluation of resulting sets.

In the following, we briefly describe the game setting, the dataset, and the used network architectures.

\subsection{Drafting in Magic: The Gathering}
\label{sec:MTG}

Collectible card games have been around for decades and are among the most played tabletop games. However, they are also among the most complex games \cite{b10}. Of course, a good player needs to be able to play the game itself, which requires an understanding and knowledge of thousands of cards. Furthermore, \textit{deck-building}, choosing a suitable set of cards to play with, is a gigantic challenge in itself and is vastly beyond the power of exhaustive computation. 
%
We abstain from explaining the complex rules \cite{b15}, as they are not necessary to understand the contribution of this work. Instead, we provide some background information about the way cards are chosen in the used dataset.

\label{sec:Drafting}
MTG is played in a variety of different styles. For this work, we consider the format of \emph{drafting} in a game with eight players. In contrast to formats where decks are constructed separately from playing, drafting features a first game phase in which players form their decks from a selection of cards, so-called \emph{packs}. Over the course of the whole draft, each player chooses a pool of $45$ cards sequentially, from which their deck is built afterward. Players get their cards by choosing from many packs as follows: Each of the eight players in a draft starts with a full pack of $15$ cards, selects a single card from it, and passes the remaining $14$ cards on to the next player. In the following rounds, players select from $14, 13, \dots$ cards until the packs are emptied. This process is repeated two more times with new decks, such that in the end, each player has selected $3 \times 15 = 45$ cards in total. 

\subsection{Data preparation and exploration}

The \textsl{DraftSim} dataset used in this research has been collected by \citet{b1} and contains 107,949 human drafts simulated on the Web.\footnote{\url{https://draftsim.com/draft-data/}} Each draft consists of 24 packs of 15 cards distributed as explained in Section~\ref{sec:Drafting}. The dataset includes 2,590,776 separate packs, which consist of a total of 265 different cards.

We train the network on pairs of possible cards in the context of the set of cards that are already held by the player. For each decision to choose the best card from a pack of $k$ cards, $k-1$ training examples are generated, for pairing the human-selected card with each of the $k-1$ other cards in the pack. The \textsl{DraftSim} dataset contains 217,624,680 such training examples. These examples are split 80/20 into training and test data, using the same split as in \cite{b1} to allow a direct comparison.

\begin{figure}[t]
    \includegraphics[width = 0.85\linewidth]{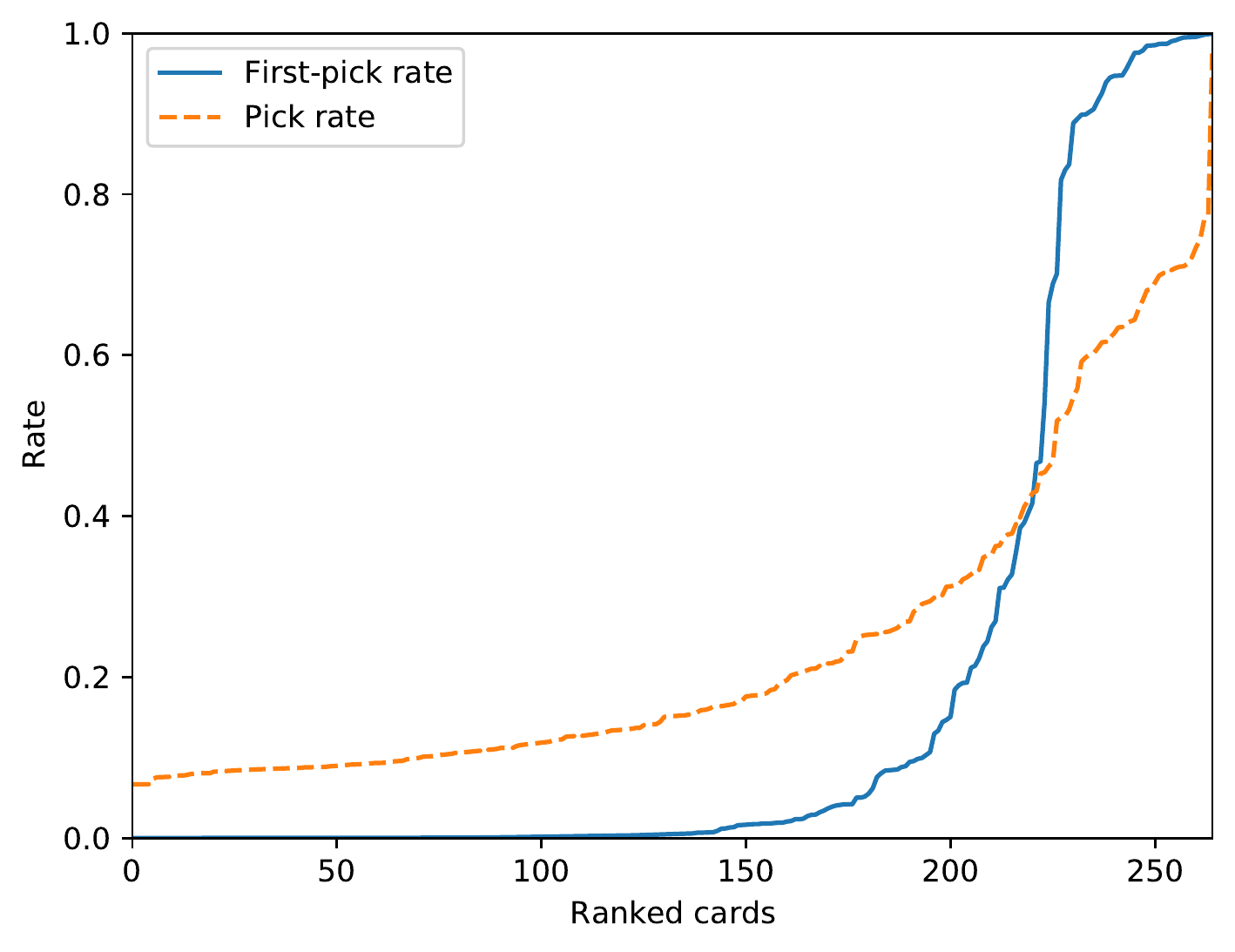}
    \caption{Pick rate per card}
    \label{fig:pickrates}
\end{figure}

To better understand the 
characteristics of the \textsl{DraftSim} dataset, we defined two metrics:
\\
\begin{enumerate}[(i)]
   \item The \emph{pick rate} of each individual card $\card$ captures how often the card was selected when being offered. $$p_{\mbox{pick}}(\card) = \frac{\textrm{number of times\ } \card \textrm{\ chosen}}{\textrm{number of times\ } \card \textrm{\ offered}}$$
    \item The \emph{first-pick rate} captures how often a card $\card$ was selected on the very first pick. $$p_{\mbox{firstPick}}(\card) = \frac{\mbox{number of times\ } \card \mbox{\ chosen first}}{\mbox{number of times\ } \card \mbox{\ offered first}}$$
\end{enumerate}

The former metric defines how likely a card is to be chosen over the whole range of the draft, while the second only considers the very first pick. Whether a card is selected first depends mainly on its individual card strength. In contrast, later card choices
are heavily influenced by previously selected cards.

Figure \ref{fig:pickrates} demonstrates that recognizing the first pick is a much easier task than choosing cards later since the consensus is higher at that point. For the first pick decision, it is possible to simply consult a ranking of available cards \cite{b4,b5,b6}. However, even for this seemingly simple task, rankings are rarely unanimous, which underlines the complexity of the domain.

Over the whole draft, all cards will be chosen at some point. For the first pick, the number of reasonable choices is relatively small, as can be seen from the quick drop of the blue solid line in Figure~\ref{fig:pickrates}. In addition, the \textbf{lowest} observed pick rate in the \textsl{DraftSim} set is $0.07$, which is close to the theoretical minimum of $\nicefrac{1}{15} \approx 0.0667$. However, the lowest first-pick rate in the data set is $0.00001$, which can safely be regarded as a misclick or otherwise unexplainable decision. In contrast, the two \textbf{highest} pick-rates 0.98 and 0.77. The best card is colorless and therefore playable in any deck. However, the second-best card is white, which explains why a portion of decisions did not choose that card, as the player was likely already firmly drafting a deck that did not include white cards. This again confirms the importance of context.


\subsection{Network Architecture}

\begin{figure}[b]
    \centering\includegraphics[width = 0.8\linewidth]{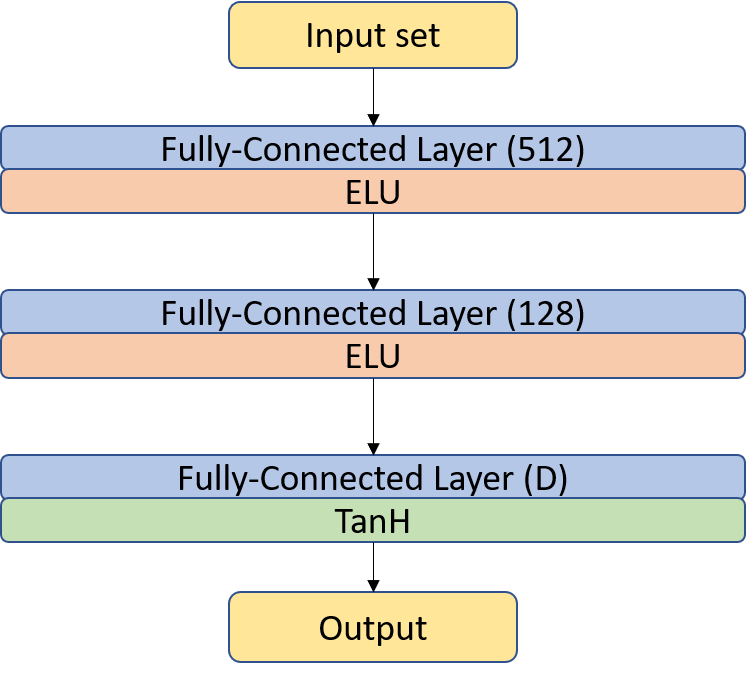}
    \caption{Siamese network \textit{N} architecture}
    \label{fig:siamese}
\end{figure}

This section shows details of the architecture and training method for the Siamese networks used in the experiments. The Siamese network encodes a set of input cards through multiple fully-connected network layers (Figure \ref{fig:siamese}). Therefore, each training update consists of two or three sequential forward passes through the network, followed by the computation of the loss and a backward pass for updating the network parameters.  The way this network is used for the twin network (RankNet) and the triplet network (CPR) is shown in Figures~\ref{fig:cross_entropy} and~\ref{fig:scheme}  respectively.

The network takes a set of cards as input. The input space is 265-dimensional, with one dimension for each possible card. For $\vect{p}$ and $\vect{n}$, the input is a one-hot encoding, while the anchor $\vect{a}$ uses an encoding in which each dimension encodes the number of already chosen cards of each type.
The output of the network is a $D$-dimensional vector of real numbers in the range $[-1,1]$, where $D \geq 1$ is a hyperparameter ($D = 1$ for RankNet). This output vector is the learned embedding of the input set.

Fully-connected layers are linked by exponential linear unit functions (ELU) \cite{b3}. In preliminary experiments, this led to quicker training than rectified linear (RELU) and leaky RELU activations. We use a learning rate of $0.0001$ and the Adam optimizer with a batch size of $128$. For the output layer, the $\tanh{}$ function was chosen. Batch normalization was not used as it did not seem to help, but we do use a dropout of $0.5$. Most of the parameters, such as the learning rate, the size of the network, and the optimizer, were not optimized, as reaching the absolute highest performance was not the priority of this work. Rather, we used intuitive parameters, which were comparable to the ones used in previous research \cite{b1}.

\section{Results}
\label{sec:results}
In this section, we discuss the performance of our networks for the card selection task and visualize the obtained card embeddings.

\subsection{Card Selection Accuracy}

Our primary goal was to compare Contextual Preference Ranking with RankNet and with previous methods for this task \cite{b1}. The best performing algorithm of that study used a traditional, single-branch deep neural network to learn a ranking over all possible cards for a given context. It was trained by directly mapping a feature-based encoding of the current set of cards $\mycards$ to a one-hot encoded vector that represents the selected card. Thus, it generated exactly one training example per card pick.
Our two agents, \textsc{CPRBot} and \textsc{RankNetBot}, instead learn on pairwise comparisons between the picked card and any other card in the candidate pack $\pack$, and therefore generate $2$ to $14$ training examples from a single pick. However, this additional constant factor in the training complexity is to some extent compensated by the fact that we were able to train our networks with a much smaller amount of training epochs.

\begin{table}[b]
\caption{Performance of proposed agent to previously seen.\\ $^1$ Heuristic agents \cite{b1} $^2$ Trained agents \cite{b1} $^3$ This work}
\medskip
\centering
\resizebox{0.7\linewidth}{!}{%
\begin{tabular}{lcl}
\toprule
\textbf{Agent} & \textbf{MTTA (\%)} & \textbf{MTPD} \\ 
\midrule
RandomBot$^1$    & 22.15 & NA     \\ 
RaredraftBot$^1$ & 30.53 & 2.62   \\ 
DraftsimBot$^1$  & 44.54 & 1.62   \\ \midrule 
BayesBot$^2$     & 43.35 & 1.74   \\ 
NNetBot$^2$      & 48.67 & 1.48   \\ \midrule 
RankNetBot$^3$      & \textbf{69.09} & \textbf{0.80}\\ 
CPRBot$^3$, D=2   & \textbf{53.69} & \textbf{0.98} \\ 
CPRBot$^3$, D=256   & \textbf{83.78} & \textbf{0.25} 
\\ \bottomrule 
\end{tabular}%
}
\label{tab:performance_comp}
\end{table}

Following \citet{b1}, Table~\ref{tab:performance_comp}  reports two measures: the \textit{mean testing top-one accuracy (MTTA)} is the percentage of cases in which the network chooses the correct card in the pack. The \emph{mean testing pick distance (MTPD)} shows how far away the correct pick is from the chosen card when ranking all possible choices. 
For \textsc{CPRBot}, we report the performance for two different agents, which only differ in the output dimension of the neural network. \textsc{RankNetBot} is able to achieve higher accuracies than the previously proposed agents, but the 256-dimensional \textsc{CPRBot} was able to achieve the best performance by a large margin.
This strong increase suggests that the Contextual Preference Ranking approach works well for this domain, and outperforms the direct comparison between two sets by RankNet. 


\begin{figure}
    \centering
    \includegraphics[width = 0.85 \linewidth]{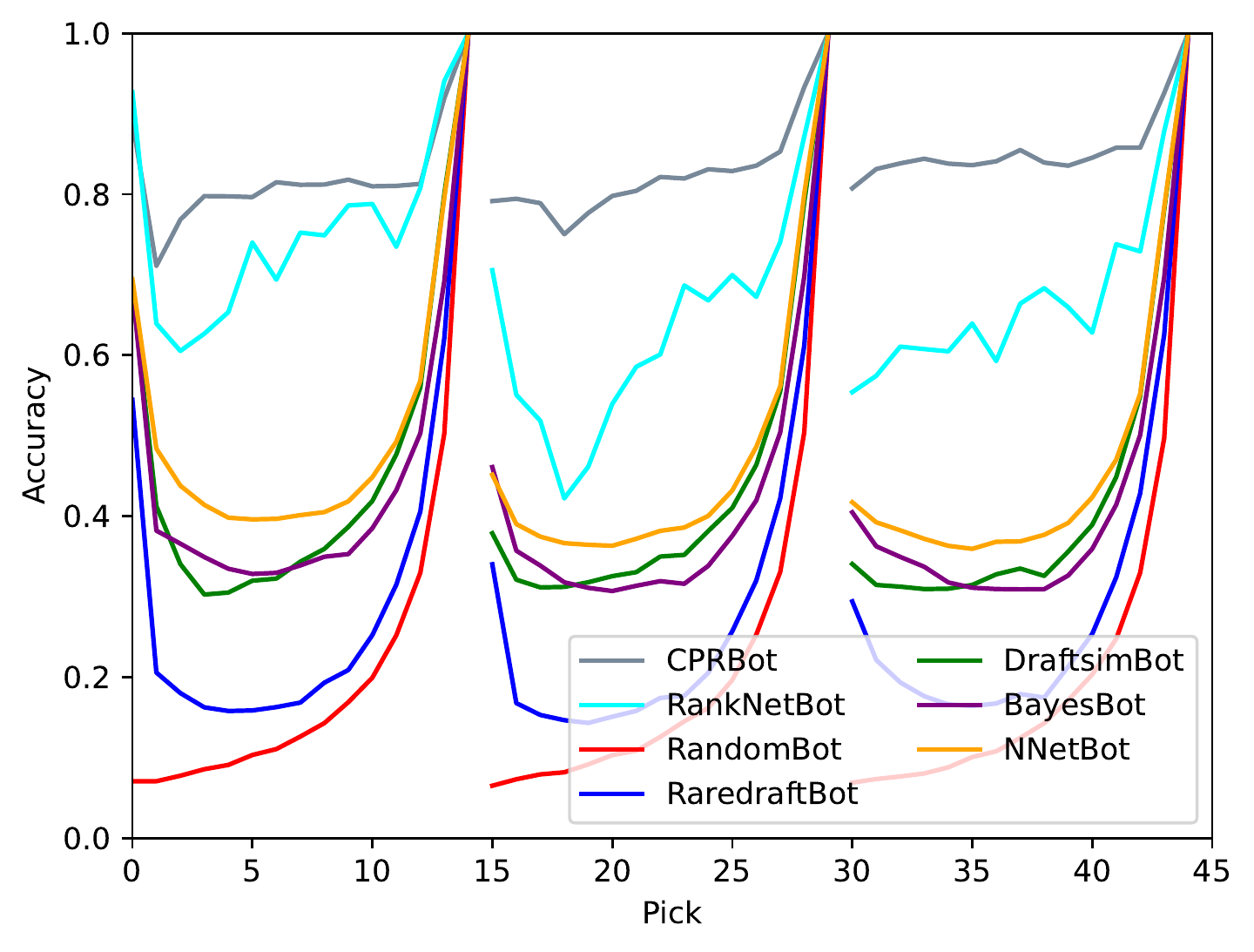}
    \caption{Accuracy of pick-prediction for size of context. Separate lines model the three separate packs.}
    \label{fig:accuracy_per_pick}
\end{figure}

\subsection{Draft Analysis}
We compare the performance of both methods over the course of the whole draft. Since already chosen cards strongly influence the current decision, we explore whether a growing set of chosen cards influences the accuracy of picks.  Figure~\ref{fig:accuracy_per_pick} shows the accuracy of all agents over the three consecutive picking rounds. Clearly, both preference-based algorithms achieve higher performance than the others. Interestingly, the accuracy of picks does not show the same performance curve for \textsc{CPRBot} as for the other methods. Those methods have U-shaped curves and fail to make good decisions in the middle of packs. \textsc{CPRBot} remains stable throughout the picks while increasing in the end due to a smaller number of choices.
\textsc{CPRBot}'s performance has an outlier at the second pick. This may be because there is no difference between the embeddings of the anchors and the embedding of the card choice. For the second pick only, the anchor set is modeled the same as a single card.

\subsection{Discussion}

\begin{figure}[t]
    \centering
    \includegraphics[width = 0.95 \linewidth]{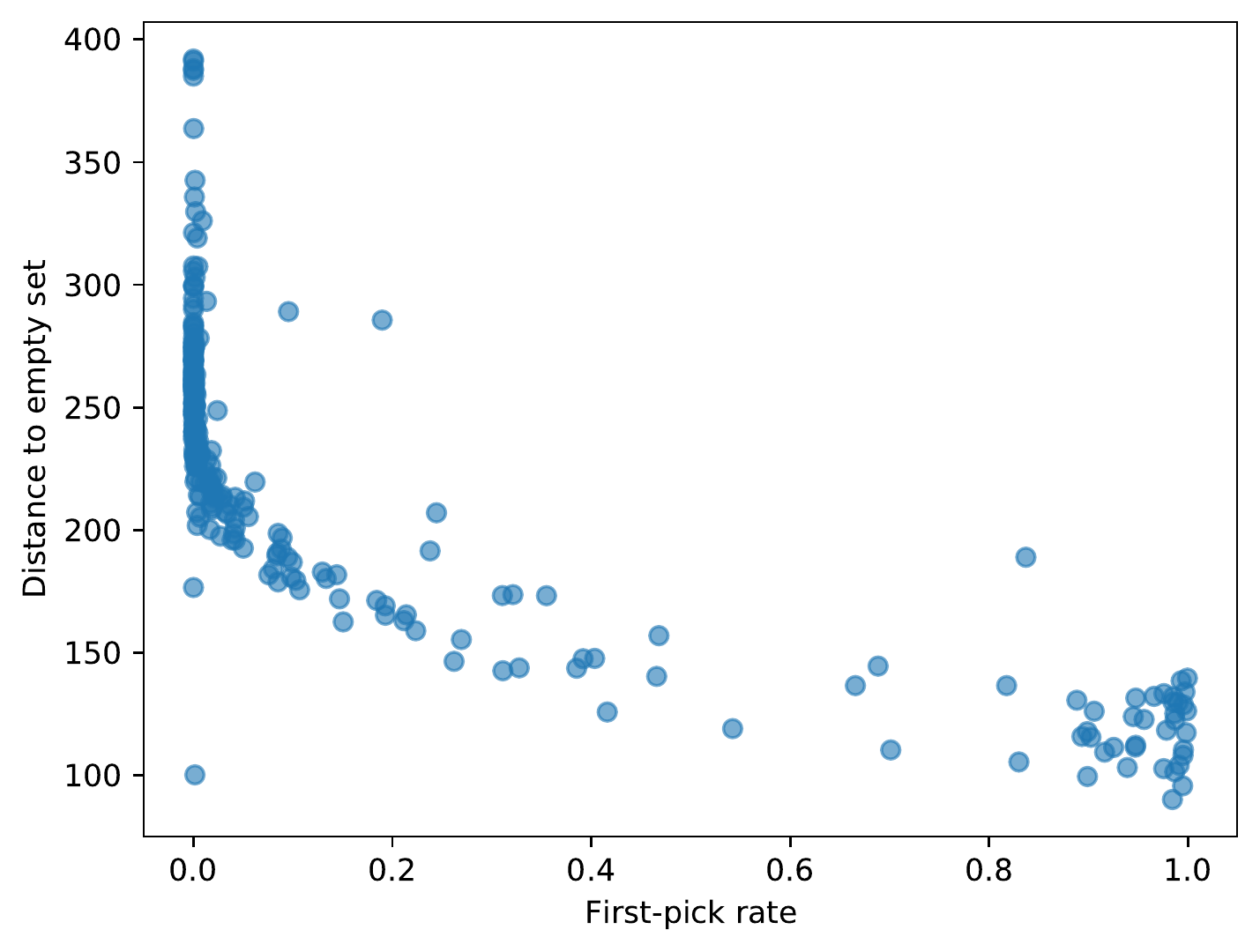}
    \caption{Correlation between FPR and distance to empty anchor. Kendall's tau = 0.74.}
    \label{fig:fpr_distance}
\end{figure}

Both preference-based agents achieved higher accuracy than previously reported. This is especially well pronounced in the middle of the pack, where weaker cards have to be compared against each other. 
To further visualize correlations between the network predictions and the underlying data, we plot the first-pick rate of cards against the distance to the empty set in Figure~\ref{fig:fpr_distance}, showing a strong correlation with a Kendall rank correlation coefficient of 0.74.
The main difference between these two statistics is that the distance is much smoother than the first-pick rate, which decreases rapidly for weaker cards. The first-pick rate is only subject to binary choices, i.e., $\card_1 \pref \card_2$\, without giving any weight to how close the decision between those cards was. Due to the training with decisions between mediocre cards, the embedding distance is a smoother measure of how strong the card is according to the network. 

\section{Conclusions and Future Work}
We showed that the Siamese-based agents which model preferences worked well in the context of drafting cards in \emph{Magic: The Gathering} and vastly outperformed previous results. Compared to \cite{b1}, we report an increase in accuracy by more than 56\%, while also decreasing the pick distance by more than 83\% with our \textsc{CPRBot}. Even when the Siamese network makes an incorrect choice, it typically ranks the correct choice very high. In addition, we also showed that the CPR-based triplet network architecture outperformed a more conventional twin network. We, therefore, speculate that the former is better suited for modeling set-addition problems with Contextual Preference Ranking.

We want to reemphasize that while these are early tests, there is no reason to believe that the success is limited to this particular setting. We did not incorporate any domain information beyond the ID of cards used to encode the input into our networks. Therefore, we expect that our proposed framework may work well for other problems where preference has to be modeled in a context. In order to further test the generality of this approach in other domains, more work with other datasets is required. We could also aim to evaluate whole sets of extensions at once instead of only single items. In addition, there is potential to use this method not only for pre-game decision-making but also for game playing for games that can be represented as sets.

One concern with our work so far is that we have only trained on human expert examples, which limits performance in a general context. For domains where an automatic evaluation of the chosen sets is available, we could generate datasets using self-play as part of an agent training loop. This is the approach used in Tesauro's TDGammon \citep{lig*Tesauro95}, in DeepMind's AlphaZero \cite{silver2018general}, and in many similar approaches.

\bibliographystyle{icml2021}
\bibliography{bibliography,jf,lig}

\end{document}